%% file: ranking_optimizers.tex
\documentclass{article}

\usepackage{times}
\usepackage{graphicx} 
\usepackage{subfigure}
\usepackage{multirow}

\usepackage{natbib}

\usepackage{algorithm}
\usepackage{algorithmic}

\usepackage{hyperref}
\usepackage{amsmath}
\usepackage{amssymb}
\include{macros}
\graphicspath{{../../figures/}}

\usepackage[accepted]{icml2016}

\icmltitlerunning{A Stratified Analysis of Bayesian Optimization Methods}

\begin{document} 

\twocolumn[
\icmltitle{A Stratified Analysis of Bayesian Optimization Methods }

\icmlauthor{Ian Dewancker}{ian@sigopt.com}
\icmlauthor{Michael McCourt}{mike@sigopt.com}
\icmlauthor{Scott Clark}{scott@sigopt.com}
\icmlauthor{Patrick Hayes}{patrick@sigopt.com}
\icmlauthor{Alexandra Johnson}{alexandra@sigopt.com}
\icmlauthor{George Ke}{l2ke@uwaterloo.ca}
\icmladdress{SigOpt, San Francisco, CA 94108 USA}

\icmlkeywords{Bayesian optimization, efficient global optimization, sequential model-based optimization, Gaussian processes, empirical analysis, evaluation methods}

\vskip 0.3in
]

\begin{abstract} 
Empirical analysis serves as an important complement to theoretical analysis for studying practical Bayesian optimization.
Often empirical insights expose strengths and weaknesses inaccessible to theoretical analysis.  We define two metrics for comparing the performance of Bayesian optimization methods and propose a ranking mechanism for summarizing performance within various genres or strata of test functions.
These test functions serve to mimic the complexity of hyperparameter optimization problems, the most prominent application of Bayesian optimization, but with a closed form which allows for rapid evaluation and more predictable behavior.
This offers a flexible and efficient way to investigate functions with specific properties of interest, such as oscillatory behavior or an optimum on the domain boundary. 
\end{abstract} 

\section{Introduction\label{sec:submission}}

Bayesian optimization is a powerful tool for optimizing objective functions which are very costly or slow to evaluate \cite{martinez2007active, brochu2010bayesian, snoek2012practical}.  In particular, we consider problems where the maximum is sought for an expensive function $f : \cX \to \RR$,

\begin{equation*}
\xx_{opt} = \argmax_{\xx \in \cX} f(\xx),
\end{equation*}

within a domain $\cX\subset\RR^d$ which is a bounding box
(tensor product of bounded and connected univariate domains).
Numerous strategies for modeling $f$ in the Bayesian optimization setting have been suggested, including the use of Gaussian processes \cite{snoek2012practical,martinez2014bayesopt},
random forests \cite{hutter2011sequential},
and tree-structured Parzen estimators \cite{bergstra2011algorithms,bergstra2013making}.

Given the variety of alternatives, performance comparisons of various optimization methods is paramount to both researchers and practitioners.
Related work summarizing the performance of Bayesian optimization methods includes
\cite{eggensperger2013towards, martinez2014bayesopt, eggensperger2015efficient}.
Much of the literature involves the use of potentially inappropriate statistical analysis, and provides little
guidance as to how performance on multiple functions $f$ can be analyzed in chorus.  Consequently,
results often read in the form of a confidence interval (derived from a small sample size)
relevant to only a single function and without any means for broader interpretation.

To address these difficulties, \secref{sec:metrics} details a non-parametric strategy for ranking performance
between various optimizers
on multiple metrics and aggregating that performance across multiple functions.  By allowing for multiple
metrics, optimizers can be studied in more detail, e.g., considering both the quality of the solution and
the speed with which it is attained.
Our strategy permits
ties during the ranking component, allowing for low significance values and non-parametric hypothesis
tests which are less powerful but do not rely on the central limit theorem.  An example of the ranking
and aggregation process is provided as well as supplementary statistical analysis which motivates the need for
non-parametric statistics.

We demonstrate our evaluation framework on a collection of Bayesian optimization methods using an open source suite of benchmark functions.  
The use of such artificial test functions is well established within the Bayesian optimization community and plays an important role in current research \cite{snoek-etal-2015a, hernandez2015predictive, gonzales2015}.  The test functions are stratified by specific attributes, such as being unimodal or non-smooth,
as discussed in \secref{sec:benchmarkfunctions}.  \secref{sec:experimentalresults} outlines our results and provides some interpretation.
We also provide an implementation of these functions \cite{mccourt_test}, hopefully facilitating future empirical insights.

\section{Evaluating Optimization Performance\label{sec:metrics}}

Many metrics exist for describing the quality of an optimizer, and each application values them differently.
Here, we consider only two metrics, which is sufficient to demonstrate the hierarchical nature of our ranking
algorithm, but others could be included if desired.
These metrics are derived from the 
best seen traces, $\fbest[i]$, which record the best seen objective value after $i$ objective function evaluations;
a sample of $\fbest[i]$ for two different optimization methods tested 30 times
is shown in \figref{fig:test}.  The shaded
region represents the inter-quartile range and reminds us that these optimization strategies are inherently
stochastic and that any results should be interpreted statistically.

\subsection{Metrics Considered\label{sec:metricsconsidered}}

The simplest comparison to make is between the $\fbest$ values observed by each strategy after observing
the maximum number of function evaluations, denoted here with $T$.
This \textbf{Best Found} metric, $\fbest[T]$, is the first we consider when comparing results.

\begin{figure}[H]
	\centering
	\includegraphics[width=\columnwidth]{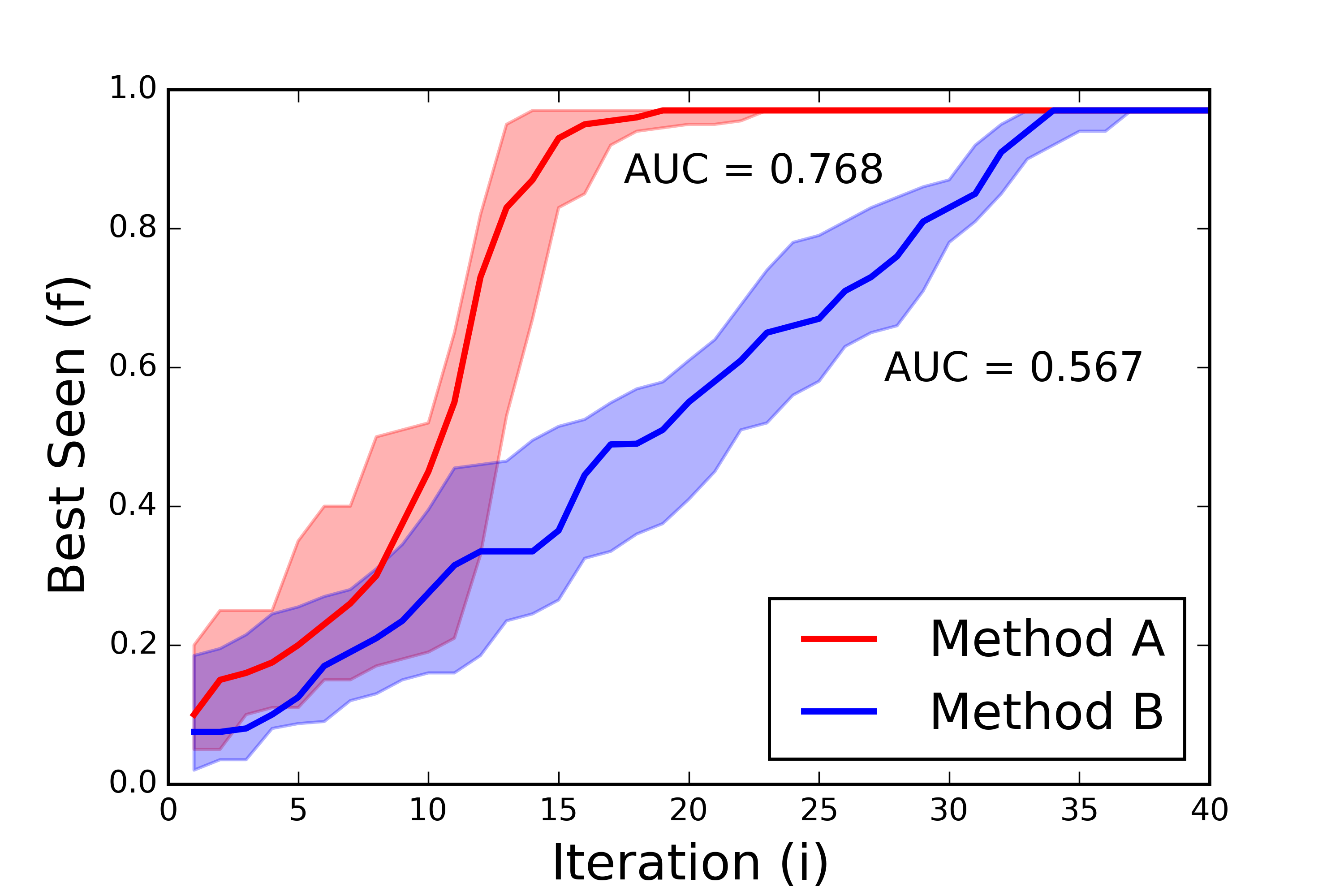}
	\caption{Hypothetical optimization methods A and B both achieve the same best found of 0.97 at the end of 40 evaluations.  Method A however finds the optimum in fewer evaluations, thus the higher AUC value.\label{fig:test}}
\end{figure}

Because each evaluation of $f$ is assumed to be expensive, we also study how quickly these methods improve.
To do so, we supplement comparisons based on only the best found metric with the \textbf{Area Under Curve} metric,
$\frac{1}{T}\sum_{i=1}^T (\fbest[i] - f_{\text{LB}})$.
A specified lower bound $ f_{\text{LB}}$ on the function ensures the AUC is always positive.   The name AUC reflects the physical interpretation of the metric as an approximate integral of the best seen traces.
\figref{fig:test} depicts two best seen traces having different AUC values.
\vspace{-1mm}

\subsection{Performance Ranking and Aggregation\label{sec:performanceranking}}

We proceed by using our metrics to establish a partial ranking (allowing for ties) of how multiple optimization
methods perform on a given function.
Because the foremost goal of optimization is to achieve  
good \textbf{Best Found} values, we reserve comparisons of \textbf{Area Under Curve} for when methods are seen as comparable with respect to the \textbf{Best Found} metric. 
Specifically, we combine the two metrics in a hierarchical fashion:
\begin{itemize}\setlength\itemsep{.1em}
	\item First, we use pairwise Mann-Whitney $U$ tests at $\alpha=.0005$ significance on the \textbf{Best Found}
		results to determine a partial ranking based only on that statistic,
	\item Any tied results from that step are then subject to additional partial ranking using the same test on
		the \textbf{Area Under Curve} metric,
	\item Ties remaining after ranking attempts using both metrics are left as ties.
\end{itemize}
This process is carried out on each function in the test suite which, in effect, allows each function to
``cast a ballot'' listing the optimization methods in order of performance on that function.  These ballots are then
aggregated using a Borda ranking system \cite{dwork2001rank} and presented in tables
in \secref{sec:experimentalresults}
alongside simpler aggregation schemes:
the number of first place finishes, and the number of at least third place finishes.
This approach generalizes to using additional metrics, provided they are applied in a specified order of importance.

As is always the case during the compilation of pairwise statistical tests, we must consider
the ``family-wise error'' $\alpha_F$, the combined probability of any type I errors given that each
test has some probability of a type I error \cite{demvsar2006statistical}.
These tests are not independent (the same samples are used for multiple tests,) but even if they were,
the probability of at least a single type I error is bounded by
\[
	\alpha_F \leq 1 - (1- \alpha)^{\binom{m}{2}},
\]
where $m$ is the number of algorithms under comparison.
In \secref{sec:experimentalresults} we use $m=7$ algorithms,
thus $\alpha_F=.01$ can be achieved with $\alpha\approx.0005$.

\subsubsection{Example Ranking and Aggregation}

Suppose we use \textsc{Method A}, \textsc{Method B}, \textsc{Method C} and \textsc{Method D} to
optimize a single function 30 times and record the \textbf{Best Found} and
\textbf{Area Under Curve} values for each optimization.
The Mann-Whitney $U$ tests on the best found value may yield the statistically significant results:
\begin{align*}
	\textsc{Method A} &> \textsc{Method D}, \\
	\textsc{Method A} &> \textsc{Method C}, \\
	\textsc{Method B} &> \textsc{Method D}.
\end{align*}
By reverse sorting these results in order of number of losses we observe the partial ranking, and
resulting Borda values,
\begin{align*}
	(\textsc{Method A},\textsc{Method B}) \!>\! \textsc{Method C} \!>\! \textsc{Method D}. \\
	2\qquad\qquad\qquad\qquad 1\qquad\qquad\quad\;\; 0\qquad\;\;
\end{align*}
Note that this is just a simple way to isolate the worst performers and certainly not
the only mechanism of creating a ranking \cite{cook2007creating}, e.g., one could rank by number of wins.
Any group of ties, such as the group with two wins in this example $(\textsc{Method A},\,\textsc{Method B})$,
is refined by studying the area under curve statistical test;
if that test stated that \textsc{Method A} had a larger value than \textsc{Method B}, that information would be
present in the final ranking
\begin{align*}
	\textsc{Method A} \!>\! \textsc{Method B} \!>\! \textsc{Method C} \!>\! \textsc{Method D}. \\
	3\qquad\qquad\quad\; 2\qquad\qquad\quad\; 1\qquad\qquad\quad\;\; 0\qquad\;\;
\end{align*}
If, on the other hand, the area under curve test was statistically insignificant, the original ranking 
and associated Borda values would be used.

If our benchmark suite consisted of only 6 functions, $f_{1 : 6}$, and each reported the following rankings:
\begin{align*}
	f_1:\quad& \textsc{A} > \textsc{B} > \textsc{C} > \textsc{D}, \\
	f_2:\quad& (\textsc{A} ,\, \textsc{B}) > \textsc{C} ,\, \textsc{D}, \\
	f_3:\quad& \textsc{C} > \textsc{A} > \textsc{B},\, \textsc{D}, \\
	f_4:\quad& \textsc{D} > (\textsc{A} ,\, \textsc{C}) > \textsc{B}, \\
	f_5:\quad& (\textsc{A} ,\, \textsc{B} ,\, \textsc{C} ,\, \textsc{D}) > 0, \\
	f_6:\quad& \textsc{B} > (\textsc{A} ,\, \textsc{C} ,\, \textsc{D}),
\end{align*}
the proposed aggregation strategies would generate the total rankings found in \tabref{tab:exampleresults}.
Tables of this form appear in \secref{sec:experimentalresults}, where the \textsc{Top Three} criterion
is more insightful than in this example; to account for ties, all algorithms at the top of a function's ``ballot''
receive credit in the \textsc{Firsts} column, and similarly for the \textsc{Top Three} column.

\begin{table}[ht]
	\centering
	\caption{Sample aggregation of results\label{tab:exampleresults}}
	\vspace{.15cm}
	\begin{small}
		\begin{sc}
			\begin{tabular}{l||ccc}
				\hline
				\multirow{2}{*}{Algorithm} & \multirow{2}{*}{Borda} & \multirow{2}{*}{Firsts} &  Top  \\
				&	&	&  Three \\
				\hline
				\abovespace
				Method A & 8 & 3 & 6 \\
				Method B & 7 & 3 & 6 \\
				Method C & 5 & 2 & 6 \\
				Method D & 3 & 2 & 5 \\
				\hline
			\end{tabular}
		\end{sc}
	\end{small}
\end{table}

\subsubsection{Statistical Considerations\label{sec:statisticalconsiderations}}
Some previous empirical analysis of optimization methods prefers
parametric statistics using the central limit theorem \cite{eggensperger2013towards,bergstra2014preliminary}
to the non-parametric statistics we employ \cite{hutter2011sequential}.
Potential non-normality of samples of $\fbest[T]$ makes parametric
statistics on small sample sizes a treacherous endeavor.

As an example, consider optimization using the simple optimization method random search
\cite{bergstra2012random}; each suggestion $\xx_i$ is chosen
\iid from $X\sim \Unif(\cX)$.
Observed function values $y_i=f(\xx_i)$ are therefore \iid realizations of some random variable
$Y$, whose distribution is determined by $f$ and $\cX$.
After $T$ random samples, the largest
observation becomes the best found value, thus $\fbest[T] = \max\{y_1,\; \ldots,\; y_T\}$.

The max of these \iid values is the $T$th order statistic $Y_{(T)}\equiv \fbest[T]$,
whose cumulative distribution function can be determined through the CDF of the random observations.
A similar result phrased in a different context is presented in \cite{bergstra2012random}.
\begin{align*}
	F_{Y_{(T)}}\!(y) \!=\! P(Y_{(T)}<y) &= P(\max\{Y_1, \ldots, Y_T\}<y) \\
		&= P(Y_1<y, \ldots, Y_T<y) \\
		&= P(Y_1<y) \cdots P(Y_T<y) \\
		&= F_{Y_1}(y)\cdots F_{Y_T}(y) \\
		&= (F_Y(y))^T.
\end{align*}

\begin{figure}[h!]
	\centering
	\includegraphics[width=\columnwidth]{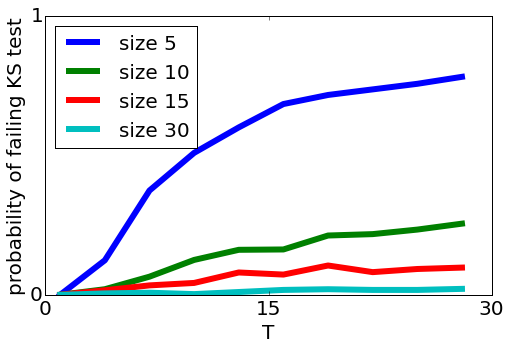}
	\caption{
		We maximized the simple 1D function $f(x) = 1-|x|$ with random search over $T$ function evaluations.
		Sample means of size $\{5, 10, 15, 30\}$ were taken from that distribution and tested with a
		Kolmogorov-Smirnov test for normality; each KS test used 500 samples and significance .05.
		We ran 800 KS tests for each $T$ value and graphed the probability of
		rejecting the null hypothesis 
		$t$-tests are appropriate.
    	\label{fig:kstest_failure}
	}
\end{figure}

The viability of a $t$-test for studying samples of $Y_{(T)}$ is dependent on how closely the
distribution of sample means matches a normal.  The central limit theorem guarantees this for sufficiently
large samples, but for small sample sizes, it seems unlikely $F_{Y_{(T)}}(y)$ matches the CDF
of a normal.  Additionally, as $T$ increases the distribution becomes increasingly skewed
negative because there is a maximum value for $Y_{(T)}$:
$f(\xx_{\text{opt}})$, the value that the optimization hopes to find.  
The impact of this on an example is
displayed graphically in \figref{fig:kstest_failure} 

\section{Benchmark Functions\label{sec:benchmarkfunctions}}

One of the prominent applications of Bayesian optimization is the efficient tuning of hyperparameters of machine learning models \cite{snoek2012practical,eggensperger2013towards}.
While hyperparameter optimization problems are some of the most important benchmarks,
it can be difficult to understand
why certain methods perform well on particular problems and others do not.

\begin{figure}[ht]
	\centering
%
	\includegraphics[width=\columnwidth]{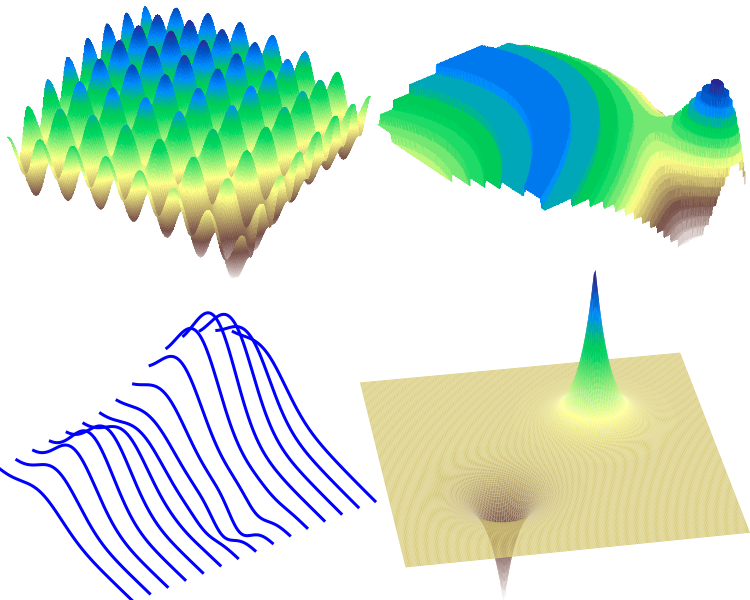}
	\caption{
		Sample benchmark functions. \textit{Top left}: Oscillatory,
		\textit{Top right}: Discrete valued,
		\textit{Bottom left}: Mixed integer,
		\textit{Bottom right}: Mostly boring,
		\label{fig:samplefunctions}
	}
\end{figure}

To try to piece together this puzzle,
we use a library of test functions having closed-form expressions which trade the
authenticity of tuning real machine learning models for speed and transparency.  
Our library, built on a core set of functions proposed by \cite{Gavana2013},
is available for inspection and acquisition \cite{mccourt_test}.
Example test functions are shown in \figref{fig:samplefunctions}.

One goal of this library is to help facilitate the transfer of insights garnered from experimentation on these
functions to real-world problems, following the path of \cite{bergstra2014preliminary,eggensperger2015efficient} which used
cheap test functions and surrogates of hyperparameter problems to evaluate Bayesian optimization software.
From a practical standpoint, their negligible evaluation time and high degree of extensibility make
artificial test functions a logical tool for testing Bayesian optimization software.

A valid criticism against using artificial test functions is that they might not be representative
of realistic problems or the objective surfaces that occur in practice.
While this statement about their limitations is perfectly accurate, test functions afford flexibility
in their design and easily allow for the presence or absence of certain properties mimicking those that also appear in
relevant circumstances, including hyperparameter optimization problems.
For example, recent work has argued that hyperparameter optimization problems may have
some level of non-stationarity \cite{snoek2014}.
We believe
that an effective way of comparing the performance of optimization methods on these types of
problems would be to develop test functions with carefully imposed non-stationarity.

To this end, we have identified characteristics which we think occur with relative frequency
in problems of interest; this list is only a sampling of characteristics discussed in this article and
not thought to be comprehensive.
\begin{itemize}
	\item \textbf{Noisy} - The function evaluation $y$ is altered by 
		$(1+\delta Z)y$ for $Z\sim\cN(0,1)$ with noise level $\delta$ between $10^{-3}$ and $10^{-1}$.
	\item \textbf{Oscillatory} - Functions $f(\xx) = S(\xx) + T(\xx)$
		with a long-range trend (possibly zero) $T$ and
		shorter scale oscillatory behavior $S$;
		the oscillations have a wavelength long enough to not act as noise.
	\item \textbf{Unimodal} - Functions with only one (or possibly zero) local maxima, with possible
		saddle points.
	\item \textbf{Boundary optima} -  $\xx_{opt}$ lies on boundary of $\cX$.
	\item \textbf{Mixed integer} - Some values in $\xx$ must be integers.  
	\item \textbf{Discrete valued} - Functions can only take finitely many values, e.g.,
		$f(\xx) = \left\lfloor \xx^T\xx \right\rfloor$ for $\xx\in[0,10]^2$ can only take the values of
		the 201 integers between 0 and 200.  This can occur in cross-validation settings.
	\item \textbf{Mostly boring} - Functions $f$ such that
		$\|f\|_{W^{1,2}(\cX)} \ll \|f\|_{W^{1,2}(\tilde{\cX})}$ where $W^{1,2}$ denotes a Sobolev space and
		$\Vol(\cX) \gg \Vol(\tilde{\cX})$ for some $\tilde{\cX}\subset\cX$.  Essentially,
		these functions have small gradient in $\cX$ except in a small region $\tilde{\cX}$.
		This is a version of non-stationarity.
	\item \textbf{Nonsmooth} - Functions which have discontinuous derivatives on manifolds of
		at least one dimension while still being in $C^0(\cX)$.
\end{itemize}

A noteworthy caveat when using artificial test functions to evaluate performance is that they
may suffer from design biases.  One example that hampered the initial iteration of this test suite involved  
having optima in predictable locations, for example, at the domain midpoint or on integer coordinates.
Under default settings, certain optimization software invokes a deterministic initialization strategy
involving $\xx$ values on the boundary or midpoint of $\cX$, which produced
unrealistically perfect results on some functions and peculiarly poor results on others.
In this benchmark suite we have made an effort to appropriately classify and segregate functions of this type, though
further work is required to identify and resolve less obvious biases.

Other work involved in the development of benchmark functions for testing optimization methods
has focused on the development of surrogate models to actual machine learning problems
\cite{eggensperger2015efficient}.  Although no such surrogates are present here, this idea
fits into the framework that we propose, and the low cost of evaluating these surrogates should
permit appropriate classification with the various attributes described above, as well as other attributes
not considered here.

\section{Ranking Demonstration\label{sec:experimentalresults}}

We conducted numerical experiments on the functions described in \secref{sec:benchmarkfunctions},
which consisted of 30--60 optimizations per algorithm per function.  The metrics described in
\secref{sec:metricsconsidered} were recorded and aggregated as described in
\secref{sec:performanceranking}.
All algorithms were terminated at 80 function evaluations, unless the function was in dimension $d<4$ when
optimization was terminated after $20d$ evaluations.  This decision was made for simplicity of comparison
and, in a real setting, each method may have a preferred stopping criteria which should be observed.

Four Bayesian optimization methods are studied in our evaluation.
\textsc{Spearmint} \cite{snoek2012practical,snoek2014,snoek_spearmint} and
\textsc{SigOpt}, which is derived from \text{MOE} \cite{Clark2014}
and all use Gaussian processes to model $f$.  We used the HIPS implementation of Spearmint \cite{snoek_spearmint}.
\textsc{SMAC} \cite{hutter2011sequential, HutterSMAC2011}
uses random forests to model $f$ and we used the python wrapper pySMAC \cite{falkner_pysmac} as implementation in our experiments.
\textsc{Hyperopt} \cite{bergstra2013making} uses
tree-structured Parzen estimators to facilitate the optimization and we used the standard python library implementation \cite{BergstraHyperopt2013} in our experiments.
In addition to these Bayesian methods, we also consider grid search (\textsc{Grid}),
random search \cite{bergstra2012random} (\textsc{Random}),
and particle swarm optimization (\textsc{PSO}) \cite{kennedy1995} as non-Bayesian baselines.  

For grid search, we chose a grid resolution per dimension so that approximately 1 million grid points were generated.  We then sampled randomly from this collection of grid points during optimization.
For \textsc{PSO}, we chose $2d$ particles in our evaluations where $d$ was the dimension of the problem.  We used an pyswarm  \cite{lee_pyswarm} as our PSO implementation in our experiments.  
We ignored the ability of the Gaussian process methods to incorporate an estimate of noise into their operation.  This was done to eliminate the unfair advantage that GP based methods would benefit from by having prior knowledge about the objective function's noise.  Prior knowledge of the objective function's process noise is also an unrealistic assumption for true black-box optimization.

When presenting results, optimization algorithms are listed alphabetically.
Tables \ref{tab:2dfunctions}--\ref{tab:10+dfunctions} analyze performance using only
the dimension of the function as classification.
The classification of functions by their attributes in tables
\ref{tab:boringfunctions}--\ref{tab:nonsmoothfunctions}, however, provides more valuable insight.
Again, the functions we used are available
in Python \cite{mccourt_test}; not all functions provided there appear in these
results.

\input{dimension_results}

Isolating the results by dimension provides some relevant insights, notably, the emergence of
\textsc{HyperOpt} into the top tier for higher dimensional functions.  It is helpful to balance the
\textsc{Borda} ranking, which rewards consistently ranking above methods,
with the \textsc{Top Three} ranking, which is more of an absolute measurement because we allow
for ties, i.e., five methods could be in the top three.

\input{experimental_results}

\begin{itemize}
	\item The GP backed algorithms \textsc{SigOpt} and \textsc{Spearmint} perform, on average,
		better on most categories.
		\begin{itemize}
			\item \textsc{Spearmint} holds slight leads in functions which are mostly boring and
			those with a boundary optimum.
			\item \textsc{SigOpt} seems to edge ahead in non-smooth and discrete functions.
		\end{itemize}
	\item \textsc{PSO} performs competitively in several function classes, including oscillatory
		and noisy, despite its non-Bayesian foundation.
	\item Noisy functions are difficult to study because of the randomness in evaluating $f$ which
		augments randomness in the Bayesian optimization.  This contributes to the broad
		representation in the \textsc{Top Three} ranking.
	\item Several methods consistently share the top three rankings for the mixed integer functions.
		That is also the case for oscillatory functions, but with a larger \textsc{Borda} gap.
	\item Mixed integer and unimodal functions have some dissonance between the \textsc{Borda}
		ranking and the \textsc{Top Three} ranking.
	\item SMAC is designed for high dimensional problems with categorical or conditional parameters,
		and none of these functions fit that description.
\end{itemize}

It should be noted that function attributes are not exclusive---a unimodal function can also be
mostly boring and have a boundary optimum.  Overlap is, however, kept to a minimum to better
focus on the impact of individual attributes.  Future experiments could involve the development of larger
groups of multiply attributed functions.

\subsection{Comparison with $t$-tests}
In \secref{sec:statisticalconsiderations}, we explained our preference for the Mann-Whitney $U$ test
over traditional $t$-tests for comparing optimization performance.  \tabref{tab:ttesttable} reinforces
the potential differences between the parametric and non-parametric tests; while the results
are not strikingly different, \textsc{Spearmint}'s performance is much clearer.  Note that all tests,
both the $t$-tests and $U$ tests, were still conducted with the significance $.0005$ explained in
\secref{sec:performanceranking}.

\begin{table}[h!]
	\centering
	\caption{\textbf{Boundary optimum functions} ($U\,/\,t$ results)\label{tab:ttesttable}}
	\vspace{.15cm}
	\begin{small}
		\begin{sc}
			\begin{tabular}{l||ccc}
				\hline
				\multirow{2}{*}{Algorithm} & \multirow{2}{*}{Borda} & \multirow{2}{*}{Firsts} &  Top  \\
				&	&	&  Three \\
				\hline
				\abovespace
				Spearmint & $74\,/\,68$ & $11\,/\,9$ & $13\,/\,13$ \\
				SigOpt    & $67\,/\,67$ & $4\,/\,8$ & $13\,/\,13$ \\
				HyperOpt   & $29\,/\,30$ & $-\,/\,-$ & $9\,/\,8$ \\
				SMAC & $7\,/\,3$ & $-\,/\,-$ & $2\,/\,1$ \\			
				\hline
				\abovespace
				PSO    & $30\,/\,33$ & $-\,/\,1$ & $11\,/\,11$ \\
				Grid    & $14\,/\,17$ & $-\,/\,1$ & $5\,/\,1$ \\			
				Random    & $2\,/\,3$ & $-\,/\,-$ & $1\,/\,-$ \\
				\hline
			\end{tabular}
		\end{sc}
	\end{small}
\end{table}

\subsection{Comparison without AUC specification}
As explained in \secref{sec:performanceranking}, the ranking scheme we propose involves a hierarchy of
metrics: first the \textbf{Best Found} and second the \textbf{Area Under Curve}.  Of course, more metrics
could be incorporated into the ranking, but two is sufficient for a demonstration.  Had we instead only
considered the best found metric we would have seen less refined tables with more ties.
\tabref{tab:noauctable} gives an example of the contrast between ranking with and without the AUC.

\begin{table}[h!]
	\centering
	\caption{\textbf{Mixed integer functions} (\textbf{AUC} / no \textbf{AUC} results)\label{tab:noauctable}}
	\vspace{.15cm}
	\begin{small}
		\begin{sc}
			\begin{tabular}{l||ccc}
				\hline
				\multirow{2}{*}{Algorithm} & \multirow{2}{*}{Borda} & \multirow{2}{*}{Firsts} &  Top  \\
				&	&	&  Three \\
				\hline
				\abovespace
				SigOpt    & $63\,/\,58$ & $10\,/\,11$ & $14\,/\,14$ \\
				Spearmint & $55\,/\,54$ & $6\,/\,10$ & $10\,/\,11$ \\
				HyperOpt   & $30\,/\,27$ & $3\,/\,3$ & $11\,/\,11$ \\		
				SMAC & $15\,/\,12$ & $2\,/\,2$ & $7\,/\,7$ \\
				
				\hline
				\abovespace
				PSO    & $32\,/\,31$ & $3\,/\,4$ & $13\,/\,13$ \\
				Grid    & $11\,/\,9$ & $2\,/\,2$ & $6\,/\,6$ \\				
				Random    & $15\,/\,14$ & $3\,/\,3$ & $7\,/\,7$ \\
				\hline
			\end{tabular}
		\end{sc}
	\end{small}
\end{table}

\section{Conclusions}

Problems such as hyperparameter optimization require efficient algorithms because of the cost
of conducting the relevant machine learning tasks.  Bayesian optimization continues to be the
standard tool for hyperparameter selection, but comparing optimization strategies is complicated
by the opaque nature of these hyperparameter optimization problems.  We have proposed an
extensible framework with which to analyze these methods, including a suite of test
functions classified by their relevant attributes and an aggregation scheme for interpreting
relative performance.

The results presented here demonstrate that
by using test functions with desired properties, e.g., non-smooth functions,
we can better decipher the performance of optimization methods.  Future
work will include identifying the corresponding attributes of hyperparameter selection problems
so that these results can be appropriately applied.  We welcome contributions and extensions to
this initial evaluation set and hope it will be useful in future empirical studies.

\bibliography{citations}
\bibliographystyle{icml2016}

\newpage
\hfill
\newpage
\input{supplementary_material}

\end{document}

%% file: macros.tex
\DeclareMathOperator*{\argmax}{arg\,max}

\DeclareMathOperator*{\Vol}{Vol}
\DeclareMathOperator*{\Unif}{Unif}

\newcommand{\iid}{\frenchspacing i.i.d. \nonfrenchspacing}

\newcommand{\fbest}{f_{\text{best}}}

\newcommand{\dif}{\textup{d}}
\newcommand{\me}{\textup{e}}

\newcommand{\xx}{{\mathbf{x}}}
\newcommand{\xxopt}{{\mathbf{x}_{\text{opt}}}}

\newcommand{\cN}{\mathcal{N}}
\newcommand{\cX}{\mathcal{X}}

\newcommand{\RR}{\mathbb{R}}

\newcommand{\figref}[1]{Figure \ref{#1}}
\newcommand{\tabref}[1]{Table \ref{#1}}
\newcommand{\secref}[1]{Section \ref{#1}}

%% file: dimension_results.tex
\begin{table}[ht!]
	\centering
	\caption{\textbf{2D functions} (13 total)\label{tab:2dfunctions}}
	\vspace{.15cm}
	\begin{small}
		\begin{sc}
			\begin{tabular}{l||ccc}
				\hline
				\multirow{2}{*}{Algorithm} & \multirow{2}{*}{Borda} & \multirow{2}{*}{Firsts} &  Top  \\
				&	&	&  Three \\
				\hline
				\abovespace
				SigOpt    & 70 & 12 & 13 \\
				Spearmint & 39 & 2 & 9 \\
				HyperOpt    & 28 & 2 & 11 \\
				SMAC & 4 & 1 & 4 \\
				
				\hline
				\abovespace
				PSO    & 31 & 1 & 12 \\
				Grid    & 8 & 1 & 6 \\
				Random    & 8 & 1 & 5 \\
				\hline
			\end{tabular}
		\end{sc}
	\end{small}
	\vspace{-.2cm}
\end{table}

\begin{table}[ht!]
	\centering
	\caption{\textbf{3--5D functions} (12 total)\label{tab:35dfunctions}}
	\vspace{.15cm}
	\begin{small}
		\begin{sc}
			\begin{tabular}{l||ccc}
				\hline
				\multirow{2}{*}{Algorithm} & \multirow{2}{*}{Borda} & \multirow{2}{*}{Firsts} &  Top  \\
				&	&	&  Three \\
				\hline
				\abovespace
				SigOpt    & 64 & 10 & 12 \\
				Spearmint & 46 & 3 & 8 \\
				HyperOpt    & 41 & 3 & 7 \\
				SMAC & 7 & -- & 1 \\			
				\hline
				\abovespace
				PSO    & 39 & 1 & 10 \\
				Grid    & 8 & 1 & 1 \\
				Random    & 6 & -- & -- \\
				\hline
			\end{tabular}
		\end{sc}
	\end{small}
	\vspace{-.2cm}
\end{table}

\begin{table}[ht!]
	\centering
	\caption{\textbf{6--9D functions} (11 total)\label{tab:69dfunctions}}
	\vspace{.15cm}
	\begin{small}
		\begin{sc}
			\begin{tabular}{l||ccc}
				\hline
				\multirow{2}{*}{Algorithm} & \multirow{2}{*}{Borda} & \multirow{2}{*}{Firsts} &  Top  \\
				&	&	&  Three \\
				\hline
				\abovespace
				SigOpt    & 58 & 10 & 11 \\
				Spearmint & 42 & 2 & 8 \\
				HyperOpt    & 36 & 2 & 11 \\
				SMAC & 10 & 1 & 1 \\
				\hline
				\abovespace
				PSO    & 34 & 2 & 9 \\
				Grid    & 6 & 1 & 1 \\
				Random    & 6 & -- & -- \\
				\hline
			\end{tabular}
		\end{sc}
	\end{small}
	\vspace{-.2cm}
\end{table}

\begin{table}[ht!]
	\centering
	\caption{\textbf{$\geq$10D functions} (8 total)\label{tab:10+dfunctions}}
	\vspace{.15cm}
	\begin{small}
		\begin{sc}
			\begin{tabular}{l||ccc}
				\hline
				\multirow{2}{*}{Algorithm} & \multirow{2}{*}{Borda} & \multirow{2}{*}{Firsts} &  Top  \\
				&	&	&  Three \\
				\hline
				\abovespace
				SigOpt    & 40 & 7 & 8 \\
				HyperOpt    & 35 & 4 & 8 \\
				Spearmint & 23 & 1 & 6 \\
				SMAC & 12 & -- & 3 \\			
				\hline
				\abovespace
				Random    & 16 & 1 & 3 \\
				PSO    & 12 & -- & 2 \\
				Grid    & -- & -- & -- \\
				\hline
			\end{tabular}
		\end{sc}
	\end{small}
	\vspace{-.2cm}
\end{table}

\eject

%% file: experimental_results.tex
\begin{table}[ht!]
	\centering
	\caption{\textbf{Mostly boring functions} (10 total)\label{tab:boringfunctions}}
	\vspace{.15cm}
	\begin{small}
		\begin{sc}
			\begin{tabular}{l||ccc}
				\hline
				\multirow{2}{*}{Algorithm} & \multirow{2}{*}{Borda} & \multirow{2}{*}{Firsts} &  Top  \\
				&	&	&  Three \\
				\hline
				\abovespace
				Spearmint & 46 & 6 & 8 \\
				SigOpt    & 45 & 5 & 10 \\
				HyperOpt    & 33 & 2 & 8 \\
				SMAC & 11 & 2 & 2 \\
				\hline
				\abovespace
				PSO    & 21 & 1 & 7 \\
				Grid    & 8 & 2 & 2 \\
				Random    & 5 & 1 & 1 \\
				\hline
			\end{tabular}
		\end{sc}
	\end{small}
	\vspace{-.2cm}
\end{table}

\begin{table}[ht!]
	\centering
	\caption{\textbf{Boundary optimum functions} (13 total)\label{tab:boundaryfunctions}}
	\vspace{.15cm}
	\begin{small}
		\begin{sc}
			\begin{tabular}{l||ccc}
				\hline
				\multirow{2}{*}{Algorithm} & \multirow{2}{*}{Borda} & \multirow{2}{*}{Firsts} &  Top  \\
				&	&	&  Three \\
				\hline
				\abovespace
				Spearmint & 74 & 11 & 13 \\
				SigOpt    & 67 & 4 & 13 \\
				HyperOpt    & 29 & -- & 9 \\
				SMAC & 7 & -- & 2 \\
				\hline
				\abovespace
				PSO    & 30 & -- & 11 \\
				Grid    & 14 & -- & 5 \\			
				Random    & 2 & -- & 1 \\
				\hline
			\end{tabular}
		\end{sc}
	\end{small}
	\vspace{-.2cm}
\end{table}

\begin{table}[ht!]
	\centering
	\caption{\textbf{Noisy functions} (24 total)\label{tab:noisyfunctions}}
	\vspace{.15cm}
	\begin{small}
		\begin{sc}
			\begin{tabular}{l||ccc}
				\hline
				\multirow{2}{*}{Algorithm} & \multirow{2}{*}{Borda} & \multirow{2}{*}{Firsts} &  Top  \\
				&	&	&  Three \\
				\hline
				\abovespace
				SigOpt    & 89 & 21 & 24 \\
				Spearmint & 60 & 9 & 16 \\
				HyperOpt    & 47 & 12 & 19 \\			
				SMAC & 13 & 4 & 6 \\			
				\hline
				\abovespace
				PSO    & 64 & 12 & 24 \\
				Grid    & 22 & 8 & 11 \\
				Random    & 16 & 7 & 9 \\
				\hline
			\end{tabular}
		\end{sc}
	\end{small}
	\vspace{-.2cm}
\end{table}

\begin{table}[ht!]
	\centering
	\caption{\textbf{Mixed integer functions} (15 total)\label{tab:integerfunctions}}
	\vspace{.15cm}
	\begin{small}
		\begin{sc}
			\begin{tabular}{l||ccc}
				\hline
				\multirow{2}{*}{Algorithm} & \multirow{2}{*}{Borda} & \multirow{2}{*}{Firsts} &  Top  \\
				&	&	&  Three \\
				\hline
				\abovespace
				SigOpt    & 63 & 10 & 14 \\
				Spearmint & 55 & 6 & 10 \\
				HyperOpt    & 30 & 3 & 11 \\		
				SMAC & 15 & 2 & 7 \\
				
				\hline
				\abovespace
				PSO    & 32 & 3 & 13 \\
				Random    & 15 & 3 & 7 \\
				Grid    & 11 & 2 & 6 \\
				\hline
			\end{tabular}
		\end{sc}
	\end{small}
	\vspace{-.2cm}
\end{table}

\eject

\begin{table}[ht!]
	\centering
	\caption{\textbf{Oscillatory functions} (11 total)\label{tab:oscillatoryfunctions}}
	\vspace{.15cm}
	\begin{small}
		\begin{sc}
			\begin{tabular}{l||ccc}
				\hline
				\multirow{2}{*}{Algorithm} & \multirow{2}{*}{Borda} & \multirow{2}{*}{Firsts} &  Top  \\
				&	&	&  Three \\
				\hline
				\abovespace
				SigOpt    & 42 & 11 & 11 \\
				SMAC & 20 & 5 & 8 \\
				HyperOpt    & 18 & 6 & 10 \\
				Spearmint & 6 & -- & 2 \\
				\hline
				\abovespace
				PSO    & 21 & 6 & 11 \\
				Random    & 11 & 3 & 7 \\
				Grid    & 9 & 4 & 7 \\
				\hline
			\end{tabular}
		\end{sc}
	\end{small}
	\vspace{-.2cm}
\end{table}

\begin{table}[ht!]
	\centering
	\caption{\textbf{Unimodal functions} (12 total)\label{tab:unimodalfunctions}}
	\vspace{.15cm}
	\begin{small}
		\begin{sc}
			\begin{tabular}{l||ccc}
				\hline
				\multirow{2}{*}{Algorithm} & \multirow{2}{*}{Borda} & \multirow{2}{*}{Firsts} &  Top  \\
				&	&	&  Three \\
				\hline
				\abovespace
				SigOpt    & 56 & 7 & 11 \\
				Spearmint & 47 & 6 & 9 \\
				HyperOpt   & 38 & 3 & 12 \\
				SMAC & 10 & 1 & 2 \\
				\hline
				\abovespace
				PSO    & 37 & 3 & 10 \\
				Random    & 8 & 1 & 1 \\
				Grid    & 7 & 1 & 1 \\
				\hline
			\end{tabular}
		\end{sc}
	\end{small}
	\vspace{-.2cm}
\end{table}

\begin{table}[ht!]
	\centering
	\caption{\textbf{Discrete functions} (12 total)\label{tab:discretefunctions}}
	\vspace{.15cm}
	\begin{small}
		\begin{sc}
			\begin{tabular}{l||ccc}
				\hline
				\multirow{2}{*}{Algorithm} & \multirow{2}{*}{Borda} & \multirow{2}{*}{Firsts} &  Top  \\
				&	&	&  Three \\
				\hline
				\abovespace
				SigOpt    & 62 & 9 & 12 \\
				Spearmint & 47 & 4 & 11 \\
				HyperOpt    & 36 & 1 & 9 \\			
				SMAC & 6 & -- & 1 \\		
				\hline
				\abovespace
				PSO    & 30 & 2 & 9  \\
				Random    & 11 & 1 & 2  \\
				Grid    & 5 & 1 & 1 \\
				\hline
			\end{tabular}
		\end{sc}
	\end{small}
	\vspace{-.2cm}
\end{table}

\begin{table}[ht!]
	\centering
	\caption{\textbf{Nonsmooth functions} (10 total)\label{tab:nonsmoothfunctions}}
	\vspace{.15cm}
	\begin{small}
		\begin{sc}
			\begin{tabular}{l||ccc}
				\hline
				\multirow{2}{*}{Algorithm} & \multirow{2}{*}{Borda} & \multirow{2}{*}{Firsts} &  Top  \\
				&	&	&  Three \\
				\hline
				\abovespace
				SigOpt    & 54 & 9 & 10 \\
				Spearmint & 47 & 6 & 9 \\
				HyperOpt    & 23 & -- & 7 \\			
				SMAC & 2 & -- & 2  \\
				
				\hline
				\abovespace
				PSO    & 30 & -- & 10 \\
				Random    & 7 & -- & 1 \\
				Grid    & 6 & -- & 2 \\
				\hline
			\end{tabular}
		\end{sc}
	\end{small}
	\vspace{-.2cm}
\end{table}

%% file: supplementary_material.tex
\section*{Supplementary Material}
In \secref{sec:statisticalconsiderations}, we alluded to the skewness of $Y_{(T)}$, the result of a random
search after $T$ function evaluations, where each observed function value $Y_i$, $1\leq i\leq T$ is a random
variable with distribution determined by the function of interest $f$ and domain $\cX$.  \figref{fig:kstest_failure}
demonstrated the impact of this empirically for the simple maximization problem involving $f(x)=1-|x|$ for $x\in[-1,1]$.
In particular, it showed that the $n$ term sample mean of samples drawn from $Y_{(T)}$ failed a Kolmogorov-Smirnov
test with greater likelihood for larger $T$ and smaller $n$.  The KS test tests whether the random variables are
normally distributed, which they must be for a hypothesis test or confidence interval invoking the central limit
theorem to be appropriate.

We can determine, at least for this simple problem, the exact nature of the Berry-Esseen inequality
\cite{korolev2010upper} governing the quality of the normal approximation.  For a random variable
$W_n=\frac{1}{n}(V_1+\ldots+V_n)$
such that $E(V_i) = 0$, $E(V_i^2)=\sigma^2$ and $E(|V_i^3|)=\rho<\infty$, the Berry-Esseen inequality says
\begin{align}
	\label{eq:berryesseeninequality}
	\left|F_{W_n}(w) - F_Z(w)\right| \leq \frac{C\rho}{\sigma^3\sqrt{n}}, \quad \text{for all } w,n,
\end{align}
where $C>0$ is a constant and $Z\sim N(0,1)$.

Our goal is to study the \iid summation $\bar{Y}_{{(T)}_n}=\frac{1}{n}\left(Y_{{(T)}_1} + \ldots + Y_{{(T)}_n}\right)$,
but to apply the Berry-Esseen inequality we will have to subtract out the mean.  In \secref{sec:statisticalconsiderations}
we showed $F_{Y_{(T)}}(y)=F_Y(y)^T$, and, for this problem, $F_Y(y)=y$ so $F_{Y_{(T)}} = y^T$;
note that, were we interested instead in a \emph{minimization} problem, this order statistics analysis would yield
$F_{Y_{(1)}}(y)=1-(1-F_Y(y))^T$.  Using this CDF, we can compute
\[
	E\left(Y_{(T)}\right) \!=\!\! \int_{\RR} y\, \dif F_{Y_{(T)}}(y) \!=\!\! \int_0^1 y T y^{T-1} \,\dif y \!=\! \frac{T}{T+1}.
\]
This gives us the random variable $V_i={Y}_{{(T)}_i} \!\!- \frac{T}{T+1}$ which satisfies $E(V_i) = 0$.
Using this, we can compute
\begin{align*}
	\sigma^2 = E(V_i^2) &= \int_0^1 \left(y-\frac{T}{T+1}\right)^2\, Ty^{T-1}\dif y \\
		&= \frac{T}{(T+1)^2(T+2)}
\end{align*}
and
\begin{align*}
	\rho \!=\! E(|V_i|^3)  = &\frac{-2T(T-1)}{(T+1)^3(T+2)(T+3)} + \\
		&12\left(\frac{T}{T+1}\right)^T \!\!\!\! \frac{T^3}{(T+1)^4(T+2)(T+3)}.
\end{align*}

\begin{figure}[ht]
	\centering
	\includegraphics[width=\columnwidth]{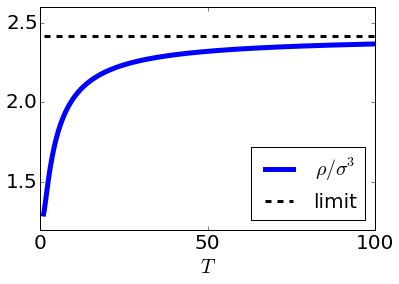}
	\caption{The $\rho/\sigma^3$ term in \eqref{eq:berryesseeninequality} is monotonically increasing with $T$, impugning
		the integrity of a $t$ test for small $n$. \label{fig:berryesseenbound}}
\end{figure}

When we finally compute this $\rho/\sigma^3$ term that governs the quality of the normal approximation
in the Berry-Esseen inequality, we see
\begin{align*}
	\frac{\rho}{\sigma^3} =-& 2\frac{(T-1)(T+2)^{1/2}}{T^{1/2}(T+3)} \\
		+&12\left(\frac{T}{T+1}\right)^T \frac{T^{3/2}(T+2)^{1/2}}{(T+1)(T+3)}.
\end{align*}
This quotient is monotonically \emph{increasing} for $T>1$, which can be determined with the derivative
(albeit with a good deal of work) or graphically as in \figref{fig:berryesseenbound}.

Of course, all is well as long as $n$ is large because $\rho/\sigma^3\to12\me^{-1} - 2$ as $T\to\infty$, but for small $n$, 
\figref{fig:berryesseenbound} demonstrates that as the quality of the solution improves ($T$ increases) the validity
of $t$ tests and central limit theorem based confidence intervals actually diminishes.  This is likely the cause of the
similar behavior for the failed Kolmogorov-Smirnov test probability in \figref{fig:kstest_failure}.